\documentclass[letterpaper]{article} 
\usepackage{aaai2026}  
\usepackage{times}  
\usepackage{helvet}  
\usepackage{courier}  
\usepackage[hyphens]{url}  
\usepackage{graphicx} 
\usepackage{natbib}  
\usepackage{caption} 
\usepackage[ruled,vlined,linesnumbered]{algorithm2e}
\usepackage{algorithmic}

\usepackage{amsmath}
\usepackage{amssymb}
\usepackage{booktabs}
\usepackage{bbding}
\usepackage{multirow}
\usepackage{multicol}
\usepackage{cuted}
\usepackage{makecell}

\usepackage{newfloat}
\usepackage{listings}
\DeclareCaptionStyle{ruled}{labelfont=normalfont,labelsep=colon,strut=off} 
\title{RadarLLM: Empowering Large Language Models to Understand Human Motion from Millimeter-Wave Point Cloud Sequence}
\author{
    Zengyuan Lai\textsuperscript{\rm 1}\equalcontrib, Jiarui Yang\textsuperscript{\rm 1}\equalcontrib, Songpengcheng Xia\textsuperscript{\rm 1}\equalcontrib, Lizhou Lin\textsuperscript{\rm 1}, Lan Sun\textsuperscript{\rm 1}, Renwen Wang\textsuperscript{\rm 2}, Jianran Liu\textsuperscript{\rm 2}, Qi Wu\textsuperscript{\rm 2}, Ling Pei\textsuperscript{\rm 1}\thanks{Corresponding Author} \\
}
\affiliations{
    \textsuperscript{\rm 1}Shanghai Jiao Tong University\\
    \textsuperscript{\rm 2}Bytedance Research\\
    \{zy.lai, jr.yang, songpengchengxia, ling.pei\}@sjtu.edu.cn,
    \{wangrenwen, liujianran\}@bytedance.com}
\begin{document}
\maketitle

\begin{figure*}[t]
  \centering
  \includegraphics[width=\linewidth]{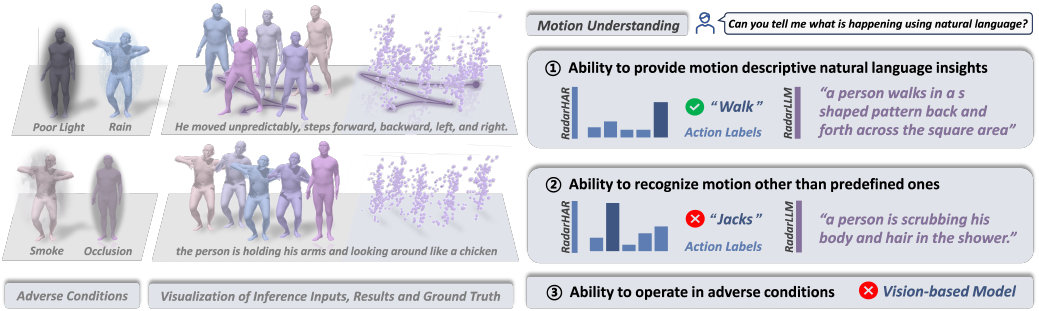}
  \caption{We propose RadarLLM, a LLM-based radar-text human motion understanding framework over traditional action label-based motion recognition in providing descriptive natural language insights, recognizing unconventional motions beyond predefined categories, and operating robustly in adverse conditions (e.g., poor lighting, occlusion, rain, and smoke).}
  \label{intro111}
\end{figure*}

\begin{abstract}
Millimeter-wave radar offers a privacy-preserving and environment-robust alternative to vision-based sensing, enabling human motion analysis in challenging conditions such as low light, occlusions, rain, or smoke. However, its sparse point clouds pose significant challenges for semantic understanding. We present RadarLLM, the first framework that leverages large language models (LLMs) for human motion understanding from radar signals. RadarLLM introduces two key innovations: (1) a motion-guided radar tokenizer based on our Aggregate VQ-VAE architecture, integrating deformable body templates and masked trajectory modeling to convert spatial-temporal radar sequences into compact semantic tokens; and (2) a radar-aware language model that establishes cross-modal alignment between radar and text in a shared embedding space. 
To overcome the scarcity of paired radar-text data, we generate a realistic radar-text dataset from motion-text datasets with a physics-aware synthesis pipeline. Extensive experiments on both synthetic and real-world benchmarks show that RadarLLM achieves state-of-the-art performance, enabling robust and interpretable motion understanding under privacy and visibility constraints, even in adverse environments.
\end{abstract}

\begin{links}
\link{Code}{https://inowlzy.github.io/RadarLLM/}
\link{Extended version}{https://arxiv.org/abs/2504.09862}
\end{links}

\section{Introduction}
\label{sec:intro}

Human motion understanding is critical in applications such as elderly care, smart home automation, and health monitoring~\cite{lai2024smart, zhang2024dynamic, xia2024envposer, shan2025mojito}. These scenarios require robust and non-intrusive sensing technologies capable of analyzing human activities while preserving privacy~\cite{xia2024timestamp, song2025predicting, xu2025ai}. However, traditional vision-based systems face significant limitations due to lighting variations, occlusions, and privacy concerns, making them unsuitable for real-world, long-term deployment.

Millimeter-wave (mmWave) radar offers a promising alternative, providing privacy-preserving motion sensing that is robust to poor lighting, occlusions, rain, and smoke, and does not capture visual identity~\cite{ding2024milliflow, gu2025hgsfusion}. Despite recent advancements in radar-based methods for activity recognition~\cite{meng2020gait, 2} and pose estimation~\cite{yang2025mmdear}, these methods primarily focus on classification or regression tasks, limiting their ability to generate fine-grained motion descriptions.

Meanwhile, large language models (LLMs) show strong capabilities in semantic reasoning across modalities like vision, audio, and motion~\cite{cho2025ambiguity, huang2024audiogpt, 43}. Inspired by this, we propose RadarLLM, the first framework bridging mmWave radar sensing and language understanding for semantic-rich motion analysis (Figure~\ref{intro111}). However, applying LLMs to radar data poses two challenges: (1) sparse, noisy point clouds hinder spatiotemporal modeling; (2) the semantic gap between radar signals and language requires sophisticated cross-modal alignment.

To address these challenges, we introduce two key components: (1) a Motion-guided Radar Tokenizer, based on an Aggregate VQ-VAE, that encodes radar point cloud sequences into discrete semantic tokens via deformable body templates and masked trajectory modeling; (2) a Radar-aware Language Model, which aligns radar tokens with textual representations in a shared embedding space to generate motion descriptions. Moreover, to overcome the lack of paired radar-text data, we propose a physics-aware synthesis pipeline that simulates realistic radar reflections from motion-text datasets, enabling effective training at scale.
In summary, our main contributions of this work are as follows:

\begin{itemize}
    \item We propose RadarLLM, the first LLM-based framework that translates low-level radar point clouds into high-level semantic motion descriptions, pioneering a new paradigm for privacy-preserving motion understanding.

    \item We introduce a novel Aggregate VQ-VAE-based Motion-Guided Radar Tokenizer. It encodes sparse radar sequences into LLM-compatible semantic tokens, leveraging deformable body templates for structural priors and masked trajectory modeling for dependency learning.

    \item We develop a physics-aware virtual radar simulator that synthesizes realistic radar-text data from motion-text datasets, effectively bypassing the bottleneck of paired real-world data scarcity for large-scale training.
\end{itemize}

\section{Related Works}
\label{sec:formatting}

\subsection{Radar-based Human Motion Understanding}
Millimeter-wave radar is robust, privacy-preserving, and effective under adverse conditions~\cite{5,7,yang2025mmdear}. Early methods extracted handcrafted Micro-Doppler features and utilized classifiers such as SVM~\cite{13} and RF~\cite{14}, achieving around 91\% accuracy on basic actions~\cite{11,10}. Deep-learning architectures, including LSTMs~\cite{17} and dual-stream CNNs~\cite{19}, automated feature learning to reach 94–99\% accuracy in controlled conditions~\cite{19,23}. Spatial-temporal transformers (ST-PCT~\cite{23}) and point-based paradigms like milliFlow~\cite{ding2024milliflow} further enhanced generalization. However, these approaches remain limited to low-level gesture classification, lacking semantic interpretation of composite activities~\cite{2,27} — a challenge that remains largely unaddressed.

\subsection{Human Motion Understanding with Multimodal LLMs}
Despite advances in sensor-based classification using RGB, IMU, and skeleton sequences~\cite{xia2024timestamp,haresamudram2025limitations,lu2025understanding}, such methods are constrained by fixed action sets and limited semantic depth. To overcome these challenges, motion-to-text frameworks have emerged, encoding discrete motion tokens for LLM translation (e.g., MotionGPT~\cite{43}, PointLLM~\cite{xu2024pointllm}, AvatarGPT~\cite{zhou2024avatargpt}) and aligning motion with natural language semantics. Building on this paradigm, Mojito extended tokenization to IMU signals~\cite{shan2025mojito}, while vision-language models have integrated visual motion cues and textual descriptions for richer interpretation~\cite{li2024mvbench}. However, radar-based LLM reasoning remains largely unexplored, a critical gap our work addresses to provide a privacy-preserving, contactless solution that eliminates the need for wearable sensors while ensuring robust perception, even in adverse scenes.

\subsection{mmWave Radar Signal Generation}
Large-scale radar–text datasets are essential for training LLM-based motion understanding models. However, existing collections ($<$9h duration, $<$40 participants, $<$27 classes)~\cite{46,47,48,49} also lack paired textual annotations~\cite{chen2022mmbody,an2022mri}. To bridge this gap, researchers have turned to synthetic data generation—but efficiency and realism remain at odds: Vid2Dopplerr~\cite{50} delivers speed at the expense of spatial detail, while RF‑Genesis~\cite{3} achieves physical fidelity only with prohibitive compute demands. These limitations have driven the emergence of our physics‐informed synthesis, which embeds motion‐capture priors to strike a balance between kinematic authenticity and scalability. At the same time, vision–language studies~\cite{51,52} and lidar-based works~\cite{an2025pre} demonstrate that such physically grounded synthetic data can closely approximate real‐world annotations. Building on these insights, we adopt SMPL-X-text pairs from HumanML3D~\cite{guo2022generating} to produce a richly annotated Radar–Text corpus via physics‐aware signal synthesis for training RadarLLM.

\section{Radar-Text Dataset Preparation}
\label{sec:virtual data generation}

\begin{figure}[ht!]
    \centering
    \includegraphics[width=\linewidth]{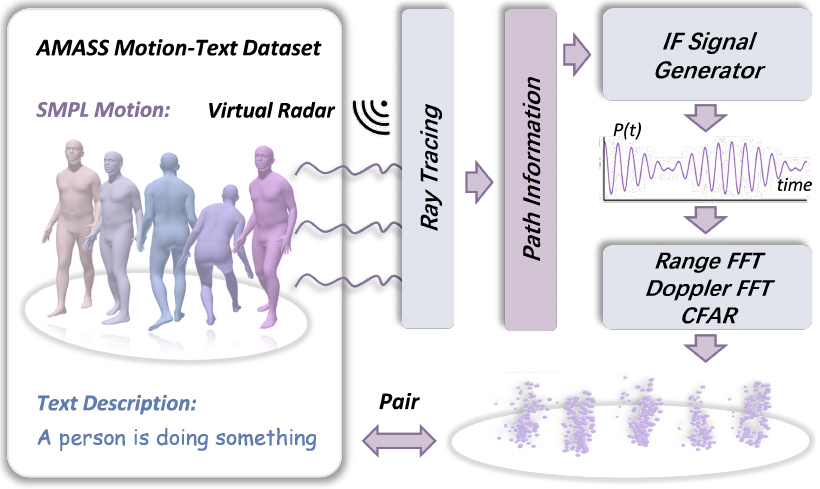}
    \caption{Virtual radar-text data generation pipeline. The Radar-Text dataset is constructed by simulating radar reflections from SMPL-X sequences using ray tracing and signal processing, based on existing motion-text datasets.}
    \label{fig:virtual data generation}
\end{figure}

\begin{figure*}[ht!]
    \centering
    \includegraphics[width=\linewidth]{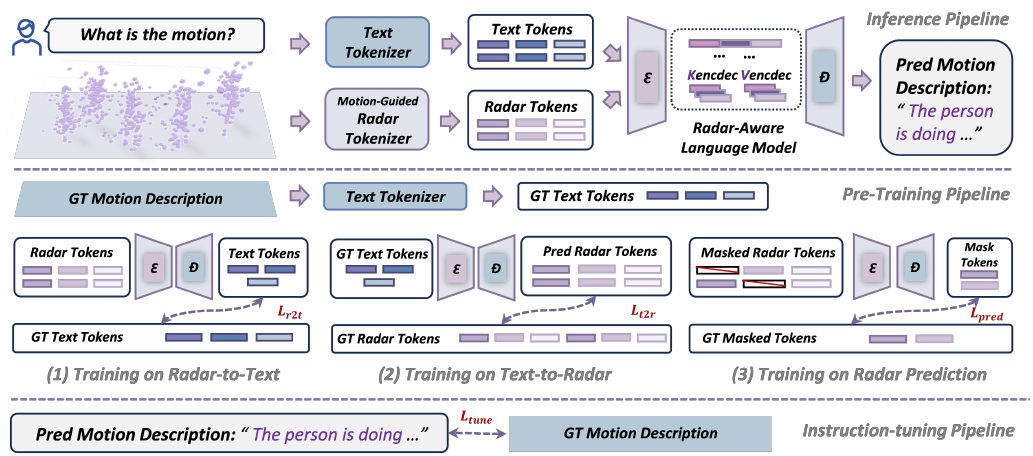}
    \caption{The overview of RadarLLM. We first encode radar point clouds into discrete tokens via a Motion-guided Radar Tokenizer. The Radar-aware Language Model then aligns these tokens with textual representations in a shared embedding space through joint optimization of unsupervised token reconstruction and supervised bidirectional radar-text translation.}
    \label{fig:arc}
\end{figure*}

\begin{figure*}[ht]
    \centering
    \includegraphics[width=\linewidth]{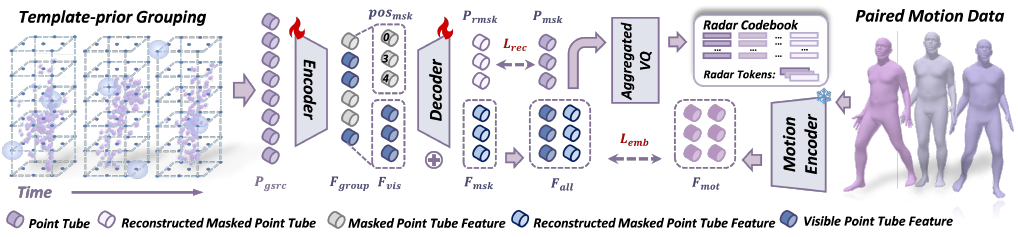}
    \caption{Architecture and training pipeline of motion-guided radar tokenizer. The Motion-guided Radar Tokenizer, built upon our Aggregate VQ-VAE architecture, compresses radar point cloud sequences into discrete semantic tokens through point cloud sequence reconstruction and motion embedding learning.}
    \label{fig:architeture}
\end{figure*}

Building an end-to-end mmWave-based motion understanding LLM is fundamentally hindered by the absence of paired radar point clouds and natural language annotations. Existing datasets with language labels focus on RGB and motion sequences, unsuitable for radar-based training. Inspired by virtual radar data generation methods~\cite{52,3}, we construct a large-scale virtual radar–text dataset from HumanML3D~\cite{guo2022generating}, comprising 13,308 SMPL-X motion sequences from AMASS~\cite{mahmood2019amass} paired with text annotations. To assess performance in real scenes, we collect a real-world test set covering one normal and four adverse environments.
\subsection{Virtual Data Preparation}
To synthesize mmWave radar signals from human motions, we employ a physics-aware synthesis. As illustrated in Figure~\ref{fig:virtual data generation}, the resulting point cloud sequences are paired with text descriptions to form our virtual radar–text dataset. 

\subsubsection{IF signal simulation}The first step is to simulate the ray paths between the rendered human meshes and the virtual radar antennas. Traditional ray tracing relies on Monte Carlo sampling, which causes a trade-off between accuracy and computational cost. So we adopt the RF adaptive sampling technique~\cite{3} to focus on the human body by edge detection. Then the path information for each ray is accumulated using the Physical Optics Integral (POI) method to obtain the simulated IF signal. 
\subsubsection{Point cloud generation}After sampling on the simulated IF signal, we first compute the Range-FFT and Doppler-FFT, followed by applying a static clutter removal algorithm to eliminate background noise by subtracting the average Doppler-FFT heatmap values across all receiving antennas. Instead of using the traditional CFAR algorithm with fixed thresholds, we directly select the 128 points per frame based on intensity from the Doppler-FFT heatmap~\cite{xue2021mmmesh}, and it ensures a consistent number of point clouds.

\subsection{Real Data Preparation} To further evaluate our method in real scenes, we collect a real‐world test set of 125 distinct motions from the HumanML3D test split, each repeated three times (375 sequences, 6–9s), recorded with a TI AWR1843BOOST radar and DCA1000EVM board under normal conditions. Another test set of four adverse conditions—rain, smoke, poor lighting, and occlusions—is created by downsampling the public mmwave dataset MMBody's~\cite{chen2022mmbody} point clouds to the 128 highest‐intensity points per frame. Point cloud synthesis follows our virtual data pipeline, and text annotations are first generated by MotionGPT conditioned on paired SMPL-X ground truth and manually checked~\cite{wang2025instructavatar}. 

\section{Method}

\subsection{Problem Statement}
Our goal is to enable a semantic-rich understanding of human motion from radar point clouds, moving beyond conventional activity classification to natural language interpretation. Given a millimeter-wave radar point cloud sequence $\mathbf{P}_{1:T}=[p_1,\dots,p_T]$, where each frame $p_t\in\mathbb{R}^{N_t\times4}$ contains the spatiotemporal coordinates $(x,y,z,t)$ of body reflections, we aim to generate a descriptive text sequence $\mathbf{Y}=[y_1,\dots,y_L]$, $y_l\in\mathcal{V}_{\text{text}}$ (predefined WordPieces vocabulary~\cite{song2020fast}). Unlike prior methods that predict discrete action labels $\hat{Y}_R$, our approach translates radar observations directly into structured textual descriptions.

To achieve this, we propose RadarLLM with three core modules: (1) constructing a virtual radar–text dataset $\{(\mathbf{P}^{syn}_{1:T},\mathbf{Y})\}$ by simulating radar reflections from human motions $\mathbf{M}_{1:T}$ via ray tracing; (2) encoding noisy sequences $\mathbf{P}^{syn}_{1:T}$ into semantic tokens $\mathbf{z}_{1:L}=\{\mathbf{z}_i\}_{i=1}^L$ through an Aggregate VQ-VAE, where $\mathbf{z}_i\in\{1,\dots,K\}$ denotes a discretized motion code, K is the total numbers of codes forming a codebook, $L=T/r$, and $r$ is the temporal downsampling rate; (3) training a conditional radar-aware language model to generate textual motion descriptions.

\subsection{RadarLLM Model Architecture}
To bridge radar point clouds and natural language, we introduce \textbf{\textit{RadarLLM}}, a unified framework that integrates mmWave radar with large language models. As shown in Figure~\ref{fig:arc}, RadarLLM comprises (1) a \textit{motion-guided radar tokenizer} that converts point cloud sequences into discrete semantic tokens, and (2) a \textit{radar-aware language model} trained via multi-modal pre-training and task-specific fine-tuning to align these tokens with textual semantics.

\subsubsection{Motion-guided Radar Tokenizer}
\label{sec:tokenizer}

To extract quantized semantic features from sparse, noisy radar point clouds for the injection into the language model, we compress spatial–temporal patterns into discrete codes via our designed \textbf{\textit{Aggregate VQ-VAE}}, leveraging human template priors and mask-enhanced spatio–temporal dependency learning, as well as motion semantics guidance. This comprises three stages: template-prior grouping, masked context aggregation, and aggregated quantization, shown in Fig~\ref{fig:architeture}. 
\\
\textbf{(1) Template-prior grouping.} To overcome the inconsistency of point location and counts among frames, we initialize $N_g$ anchors on a deterministic $N_x\times N_y\times N_z$ grid within a bounding-box template to construct temporal associations for each body region around anchors, then aggregate neighborhood points with the SOTA P4Conv encoder $\mathbf{E}$ in ~\cite{fan2021point,jing2024x4d,yang2025mmdear}, yielding $\mathbf{F}_{group}\in\mathbb{R}^{L\times N_g\times C}$.
\\
\textbf{(2) Masked Context Aggregation.} To enhance the learning of the dependencies among body parts, 50\% of anchor trajectories are masked to form visible features $\mathbf{F}_{vis}\in\mathbb{R}^{L\times N_g^{vis}\times C}$; a transformer decoder reconstructs $\mathbf{F}_{msk}=D(\mathbf{F}_{vis})$ via cross-attention, and merging with $\mathbf{F}_{vis}$ produces $\mathbf{F}_{all}=[\mathbf{F}_{vis},\mathbf{F}_{msk}]$, expected to approach motion semantic features extracted from the paired motion by a motion encoder, enriching semantics learned efficiently.
\\
\textbf{(3) Aggregated Quantization.} Aggregated quantization then maps each $\mathbf{F}_{all}^t$ to its nearest code in the trainable codebook $\mathcal{Z}=\{\mathbf{z}_k\}_{k=1}^K\subset\mathbb{R}^{512\times512}$ to construct a discrete token sequence aligned with the language model’s embedding space, enabling direct cross-modal translation, where 
\begin{equation}
\mathbf{z}_t=\arg\min_{\mathbf{z}_k\in\mathcal{Z}}\|\mathbf{F}_{all}^t-\mathbf{z}_k\|_2,\quad t=1,\dots,L.
\end{equation}
An ablation study for the effectiveness of vector quantization can be found in the supplementary file.
\\
\textbf{(4) Train and Inference Paradigm.}
In training, we optimize
\begin{equation}
\mathcal{L}_{\mathrm{VQ}}
=\mathcal{L}_{\mathrm{rec}}
+\mathcal{L}_{\mathrm{emb}}
+\mathcal{L}_{\mathrm{commit}},
\end{equation}
where
\begin{equation}
\mathcal{L}_{\mathrm{rec}}
=\frac{1}{|\mathbf{P}_{msk}|}
\sum_{\mathbf{x}\in\mathbf{P}_{msk}}
\min_{\mathbf{y}\in\mathbf{P}_{rmsk}}
\|\mathbf{x}-\mathbf{y}\|^2,
\end{equation}
forming the Chamfer Distance in~\cite{shen2023masked,shen2023pointcmp} to promote learning the spatial-temporal features of point cloud sequence via the sequence reconstruction, 
\begin{equation}
\mathcal{L}_{\mathrm{emb}}
=\|\mathbf{F}_{all}-\mathbf{F}_{mot}\|_2^2
\end{equation}
forming the motion guidance via aligning radar features with motion semantics, which is expected to accelerate feature learning. $\mathbf{F}_{mot}$ is the corresponding motion features.

\begin{equation}
\mathcal{L}_{\mathrm{commit}}
=\|\mathrm{sg}[\mathbf{F}_{all}]-\mathbf{z}\|_2^2
+\|\mathbf{F}_{all}-\mathrm{sg}[\mathbf{z}]\|_2^2.
\end{equation}
forming the commitment loss to stabilize codebook learning and enforce encoder–codebook consistency.

In inference, no masking is applied, allowing robust encoding via learned dependencies.

\subsubsection{Radar-aware Language Model}
\label{sec:llm}

To establish semantic equivalence between continuous radar patterns and discrete text and enable cross-modal reasoning, we tokenize radar clouds into discrete tokens $\mathbf{z}_{1:L}$, map them to indices $\mathbf{s}_{1:L}\in\{1,\dots,K\}^L$, and merge with word tokens in a unified vocabulary $\mathcal{V}=\mathcal{V}_{\text{text}}\cup\mathcal{V}_{\text{radar}}$ (32,768 WordPieces+$K$ radar tokens+\texttt{/som}, \texttt{/eom}). The combined input
\begin{equation}
\mathbf{X}=[w_1,\dots,w_{L_t},\,s_1,\dots,s_L]
\end{equation}
is fed into a modified T5~\cite{43}: shared embeddings project all tokens to 512 dimensions, cross-modal attention learns joint context, and the decoder auto-regressively predicts text $\mathbf{Y}=[y_1,\dots,y_L]$. 

To effectively align radar tokens with textual semantics while ensuring adaptability to diverse motion-to-text instructions, we employ a two-stage training process: a pre-training stage for robust cross-modal representation learning and an instruction-tuning stage for task-specific refinement.
\\
\textbf{(1) Pre-Training Stage.} To learn robust cross-modal representations, we adopt a multi-task training paradigm~\cite{43, zhou2024avatargpt}.

\begin{itemize}
\item \textit{Radar Prediction: } 
Following the span corruption strategy of T5~\cite{43}, we randomly mask 15\% of radar tokens and replace them with sentinel tokens. The model predicts original tokens through:
\begin{equation}
\mathcal{L}_{\text{pred}} = -\sum_{i \in \mathcal{M}} \log p(s_i | s_{\mathcal{M}}),
\end{equation}
where $\mathcal{M}$ denotes masked positions.

\item \textit{Radar→Text}: Encode radar tokens $\mathbf{z}_{1:L}$, decode text $\mathbf{w}_{1:L}$:
\begin{equation}
\mathcal{L}_{\text{r2t}} = -\sum_{t=1}^L \log p(w_t | \mathbf{z}_{1:L}, \mathbf{w}_{<t}).
\end{equation}

\item \textit{Text→Radar}: Encode text $\mathbf{w}_{1:L}$, autoregressively generate radar tokens:
\begin{equation}
\mathcal{L}_{\text{t2r}} = -\sum_{t=1}^L \log p(z_t | \mathbf{w}_{1:L}, \mathbf{z}_{<t}).
\end{equation}
\end{itemize}

The total pretraining loss combines these objectives:
{\small
\begin{equation}
\mathcal{L}_{\text{pretrain}} = \lambda_1 \mathcal{L}_{\text{pred}} + \lambda_2 \mathcal{L}_{\text{r2t}} + \lambda_3 \mathcal{L}_{\text{t2r}},
\end{equation}}
where $\lambda_1,  \lambda_2, \lambda_3$ are the hyperparameters for balancing each loss's contribution.

 The ablations on language model selection and multi-task training strategy are detailed in the Experiments section.
\\
\textbf{(2) Instruction-Tuning Stage.} To enhance task adaptability, we adopt instruction-aware prompts (e.g.\ “Describe the motion \texttt{<Motion\_Placeholder>}…”) concatenated with $\mathbf{z}$~\cite{xu2024pointllm}, and refine alignment via a similarity-based tuning loss $\mathcal{L}_{tune}$ against ground-truth descriptions.

\setlength{\tabcolsep}{1mm}
\begin{table*}[ht]
\centering
\fontsize{9}{10}\selectfont
\begin{tabular}{lccccccccc}
\toprule[1pt]
\textbf{Model} & \textbf{Data Domain} & \textbf{ROUGE-1} & \textbf{ROUGE-L} & \textbf{BLEU-1} & \textbf{BLEU-4} & \textbf{METEOR} & \textbf{CIDEr} & \textbf{BERTScore} & \textbf{SimCSE} \\
\midrule
\multirow{2}{*}{\makecell[l]{MotionGPT\textsuperscript{*} \\ \cite{43}}} & Virtual & 31.2 & 29.4 & 37.6 & 5.0 & 26.1 & 6.5 & 82.6 & 88.9 \\
 & Real & 28.0 & 25.6 & 36.1 & 2.9 & 21.9 & 3.2 & 80.5 & 87.2 \\ \midrule
\multirow{2}{*}{\makecell[l]{AvatarGPT\textsuperscript{*} \\ \cite{zhou2024avatargpt}}} & Virtual & 32.2 & 30.0 & 36.3 & 5.0 & 28.3 & 6.8 & 82.4 & 88.7 \\
 & Real & 31.0 & 28.8 & 38.1 & 4.2 & 25.6 & 5.6 & 81.4 & 88.1 \\ \midrule
\multirow{2}{*}{\makecell[l]{Video-LLaMA2\textsuperscript{*} \\ \cite{cheng2024videollama}}} & Virtual & 30.2 & 26.7 & 35.2 & 3.6 & 30.4 & 4.2 & 81.0 & 88.4 \\
 & Real & 31.4 & 28.8 & 38.3 & 4.3 & \textbf{\underline{28.6}} & \textbf{\underline{7.0}} & 80.1 & 88.0 \\ \midrule
\multirow{2}{*}{\makecell[l]{VideoChatGPT\textsuperscript{*} \\ \cite{maaz2023video}}} & Virtual & 18.6 & 16.1 & 19.5 & 0.8 & 15.3 & 1.0 & 78.5 & 85.7 \\
 & Real & 17.8 & 15.6 & 19.4 & 0.1 & 13.0 & 1.2 & 77.6 & 85.0 \\ \midrule
\multirow{2}{*}{\makecell[l]{Video-LLaVA\textsuperscript{*} \\ \cite{lin2023video}}} & Virtual & 22.8 & 19.2 & 26.7 & 1.3 & 19.2 & 2.1 & 80.3 & 87.6 \\
 & Real & 22.6 & 19.3 & 27.2 & 1.4 & 17.4 & 3.5 & 79.7 & 87.2 \\ \midrule
\multirow{2}{*}{\makecell[l]{VTimeLLM\textsuperscript{*} \\ \cite{huang2024vtimellm}}} & Virtual & 19.1 & 15.8 & 19.0 & 0.9 & 17.7 & 1.4 & 79.8 & 87.4 \\
 & Real & 21.3 & 16.9 & 24.1 & 0.8 & 18.0 & 2.1 & 79.6 & 87.3 \\ \midrule
\multirow{2}{*}{RadarLLM (Ours)} & Virtual & \textbf{38.4} & \textbf{36.0} & \textbf{48.0} & \textbf{11.4} & \textbf{33.7} & \textbf{8.3} & \textbf{83.3} & \textbf{89.6} \\
 & Real & \textbf{\underline{31.7}} & \textbf{\underline{28.8}} & \textbf{\underline{44.2}} & \textbf{\underline{5.0}} & 25.7 & 4.0 & \textbf{\underline{81.4}} & \textbf{\underline{88.1}} \\
\bottomrule[1pt]
\end{tabular}
\caption{Comparison with state-of-the-art methods on virtual and real datasets}
\label{tab:main_results}
\end{table*}

\begin{figure*}[htbp!]
    \centering
    \includegraphics[width=\linewidth]{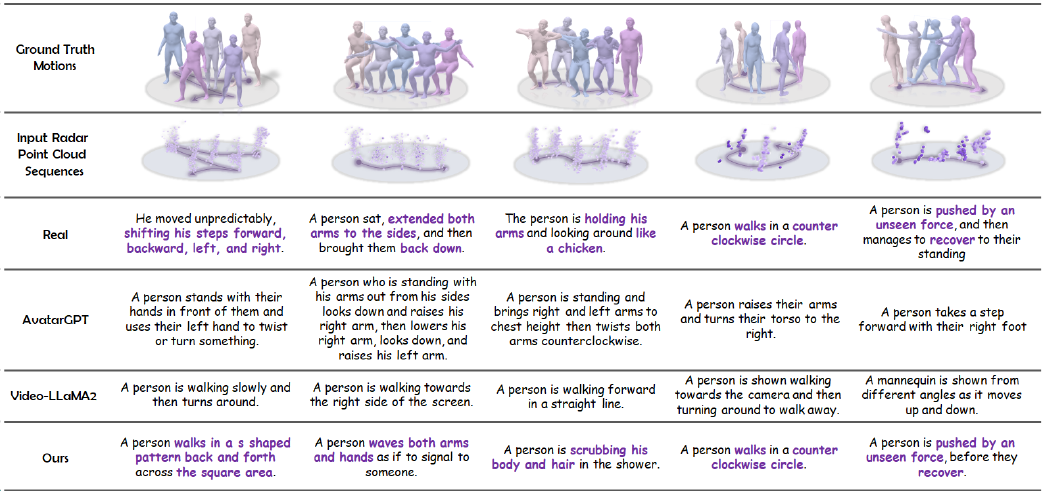}
    \caption{Visualization of predicted textual descriptions alongside corresponding motion sequences and radar point clouds. The left three columns demonstrate results on real-world normal environment data, while the right two columns showcase predictions on synthesized virtual data.}
    \label{fig:comparision_main}
\end{figure*}

\section{Experiments}
This section first introduces competing methods and metrics in our experiment (Sec.~\ref{sec:experimental setup}). We then present comprehensive comparisons on the radar-to-text task across both virtual and real datasets (Sec.~\ref{sec:comparisons on radar-to-text}). Finally, we conduct ablation studies to validate the effectiveness of each module (Sec.~\ref{sec:ablation}) and evaluate robustness in adverse environments (Sec.~\ref{sec:robust}). Additional experiment results and discussions are provided in the supplementary material.

\subsection{Experimental Setup}
\label{sec:experimental setup}
\subsubsection{Baselines} To our knowledge, RadarLLM is the first end-to-end radar-to-text framework for human motion understanding using mmWave point clouds, with no directly comparable baselines. Following the two-stage evaluation protocol of PointLLM~\cite{xu2024pointllm} and LidarLLM~\cite{yang2025lidar}, we retrain the real-time 3D human pose estimator mmMesh~\cite{xue2021mmmesh} on our radar dataset to convert sparse point clouds into SMPL‑X meshes. These SMPL‑X meshes are then rendered as videos or converted into skeleton sequences, which serve as inputs to state-of-the-art video- and motion-based text generation models, respectively, including MotionGPT~\cite{43}, AvatarGPT~\cite{zhou2024avatargpt}, Video-LLaMA2~\cite{cheng2024videollama}, Video-ChatGPT~\cite{maaz2023video}, Video-LLaVa~\cite{lin2023video}, and VTimeLLM~\cite{huang2024vtimellm}. Although these general-purpose models are originally trained on substantially larger vision and motion corpora, this unified two-stage setup ensures a fair comparison using identical inputs and underscores the strengths of our end-to-end design.
\subsubsection{Datasets} As mentioned in Sec.~\ref {sec:virtual data generation}, using HumanML3D splits (Sec.~\ref {sec:virtual data generation}), we train on virtual train-set and evaluate on virtual/real test-set. Adverse condition evaluation is tested on the mentioned down-sampled MMBody subsets.
\subsubsection{Evaluation Metrics} Following previous multi-modal text generation works~\cite{43, xu2024pointllm}, we use Rouge~\cite{lin2004rouge}, BLEU~\cite{zhang2019bertscore},  METEOR~\cite{banerjee2005meteor}, Cider~\cite{vedantam2015cider}, BertScore~\cite{zhang2019bertscore} and SimCSE~\cite{gao2021simcse} to evaluate the quality of generated captions.

\subsection{Comparisons on Radar-to-Text}
\label{sec:comparisons on radar-to-text}
To demonstrate superior cross-modal understanding performance, we conduct the radar-to-text experiments on both virtual and real data. The aforementioned radar-based HPE mmMesh and our proposed RadarLLM are trained fully on the virtual dataset, and the pre-trained motion- and video-based models are adopted to generate the text descriptions. The quantitative results on virtual and real test datasets are shown in Table~\ref{tab:main_results}. Our method achieves state-of-the-art performance across all metrics on the virtual test dataset, outperforming the strongest baseline (AvatarGPT) by +20.0\% ROUGE-L, +22.1\% CIDE, and +128\% BLEU-4 improvement, indicating better preservation of motion semantics in textual descriptions. To further evaluate the performance in real scenes, we conduct experiments on collected test data; our method remains best on almost all metrics, demonstrating the superior generalization ability to real data.

Figure~\ref{fig:comparision_main} presents qualitative comparisons of textual descriptions generated from our method and two best baselines. Our method outperforms AvatarGPT and Video-LLaMA2 by capturing fine-grained motion details and contextual semantics more accurately, while baseline methods produce generic or simple descriptions.

\subsection{Ablation Study}
\label{sec:ablation}

\setlength{\tabcolsep}{2.6mm}
\begin{table*}[t!]
\centering
\fontsize{9}{10}\selectfont
\begin{tabular}{lccccccccc}
\toprule[1pt]
Model & ROUGE-1 & ROUGE-L & BLEU-1 & BLEU-4 & METEOR & CIDEr & BERTScore & SimCSE \\
\midrule
w/o template-based anchor & 27.9& 25.7& 34.8& 3.8& 22.5& 3.2& 81.1& 87.8 \\
w/o mask for training & 35.0& 32.4& 43.1& 8.7& 31.0& \textbf{11.3}& 83.2& 89.5 \\
w/o embedding loss & 28.6& 26.5& 35.4& 4.2& 23.5& 3.8& 81.5& 88.1 \\

\midrule
\textbf{RadarLLM} & \textbf{38.4} & \textbf{36.0} & \textbf{48.0} & \textbf{11.4} & \textbf{33.7} & 8.3 & \textbf{83.3} & \textbf{89.6} \\
\bottomrule[1pt]
\end{tabular}
\caption{Ablation study on the Aggregate VQ-VAE components.}
\label{tab:ablation_vq}
\end{table*}

\setlength{\tabcolsep}{0.6mm}
\begin{table}[ht!]
\centering
\fontsize{9}{10}\selectfont
\begin{tabular}{lccccccc}
\toprule[1pt]
LLM Model & Params & FPS↑ & Self-BLEU↓ & ROUGE-L↑ & SimCSE↑ \\
\midrule
T5-small & 60M & \textbf{97.0} & \textbf{92.2} & \underline{36.0} & \underline{89.6} \\
GPT2-M & 355M & \underline{72.7} & \underline{96.2} & 35.4 & 89.5 \\
Deepseek-R1 & 1.8B & 53.6 & 98.4 & \textbf{37.4} & \textbf{89.9} \\
\bottomrule[1pt]
\end{tabular}
\caption{Ablation study on different LLM architectures.}
\label{tab:ablation_llm}
\end{table}

\setlength{\tabcolsep}{1.05mm}
\begin{table}[t]
  \centering
  \fontsize{9}{10}\selectfont
  \begin{tabular}{lcccc}
    \toprule[1pt]
    Task & ROUGE-L & BLEU-1 & METEOR & BERTScore \\
    \midrule
    R→T                & 33.0 & 42.8 & 31.2 & 82.5 \\
    R→T \& T→R         & 33.0 & 43.1 & 31.2 & 82.5 \\
    R→T \& R–Pred      & 33.9 & 43.2 & 32.4 & 82.9 \\
    \textbf{All Tasks} & \textbf{36.0} & \textbf{48.0} & \textbf{33.7} & \textbf{83.3} \\
    \bottomrule[1pt]
  \end{tabular}
  \caption{Ablation of multi-task training strategy}
  \label{1}
\end{table}

\subsubsection{Effectiveness of Aggregate VQ-VAE}
To thoroughly evaluate the architectural contributions of our radar tokenizer, we conduct a series of ablation studies focusing on three key components on the AMASS test dataset. We first replace the template-based anchor grouping mechanism with the vanilla Farthest Point Sampling (FPS), the results in Table~\ref{tab:ablation_vq} show severe performance degradation of 27.3\% in ROUGE-1, highlighting the importance of consistent spatial-semantic correspondence. We alter the training objective by reconstructing the full point cloud sequence instead of recovering the masked point tube, following the traditional self-supervised learning strategy in VAE. This change leads to a 23.7\% drop in BLEU-4 scores, demonstrating the effectiveness of our masked point tube recovery approach. Finally, by removing the embedding loss, it proves crucial for cross-modal alignment, with its absence leading to 54.2\% lower CIDEr scores.

\subsubsection{Effectiveness of LLM selection}
To assess LLM selection effectiveness and scalability across model sizes, Table~\ref{tab:ablation_llm} demonstrates that T5‑Small achieves the fastest inference and highest diversity, with only minor semantic degradation. GPT2‑Medium strikes a middle ground in speed (FPS) and diversity (Self-BLEU), albeit with somewhat lower semantic precision, while DeepSeek‑R1 delivers the best lexical and semantic scores at the expense of throughput and increased repetition. Under limited computing resources, LoRA tuning on GPT2 and DeepSeek may reinforce frequent token patterns, suggesting that full fine‑tuning under sufficient resources can be explored; Thus, T5‑Small remains the most balanced choice. 

\subsubsection{Effectiveness of Multi-task training strategy}
To evaluate multi-task objectives, we compare four schemes in Table~\ref{1}: Radar→Text (R→T), R→T + Text→Radar (T→R), R→T + radar-prediction (R–Pred), and all three tasks. Adding T→R yields the arising of  BLEU-1 (+0.7\%). Incorporating R–Pred improves ROUGE-L by 2.7\%, BLEU-1 by 0.9\%, METEOR by 3.8\%, and BERTScore by 0.5\%. Jointly optimizing all objectives boosts ROUGE-L by 9.1\%, BLEU-1 by 12.2\%, METEOR by 8.0\%, and BERTScore by 1.0\%, confirming that multi-task learning substantially enhances motion-to-text performance.

\subsection{Robustness under Adverse Environments}
\label{sec:robust}
\begin{figure}[t]
    \centering
    \includegraphics[width=\linewidth]{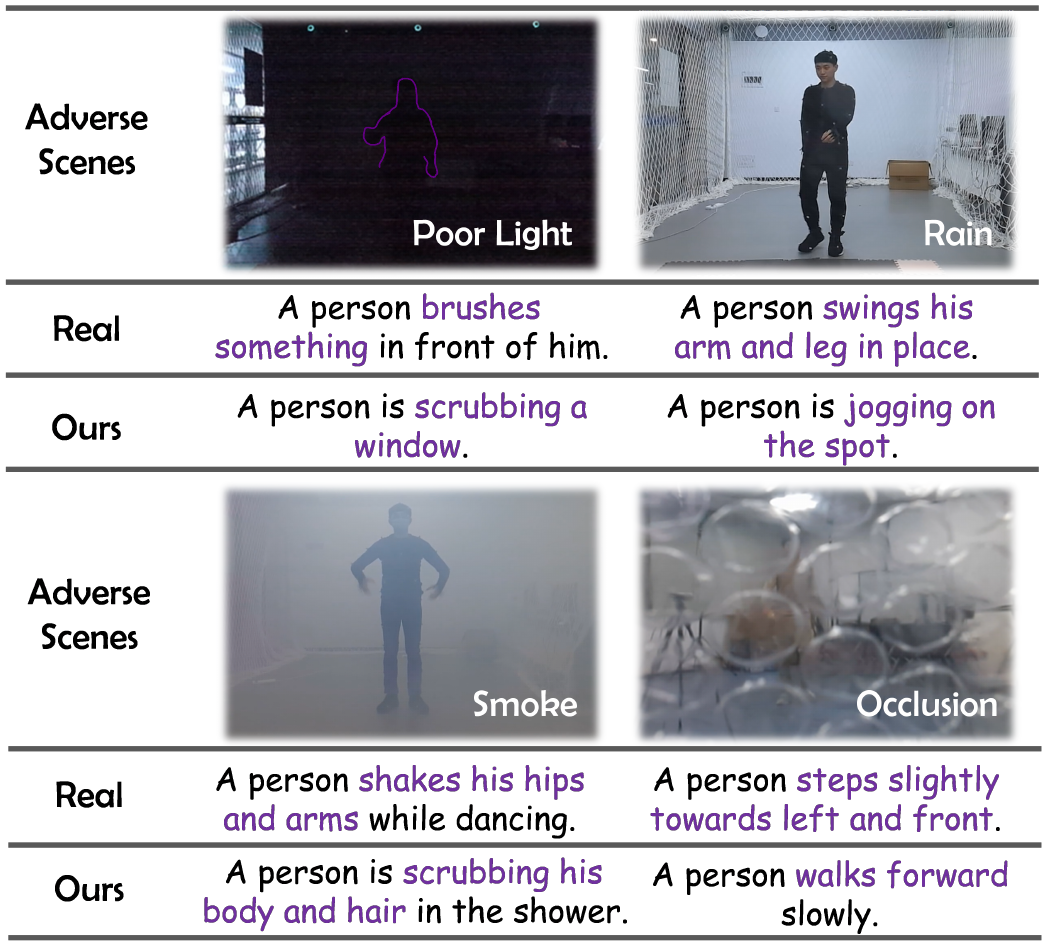}
    \caption{Visualization of predictions in adverse conditions.}
    \label{adverse}
\end{figure}

To assess the robustness of RadarLLM under adverse environmental conditions, we evaluated it on four corrupted subsets of the MMBody dataset—rain, smoke, poor lighting, and occlusions. ROUGE‑L declines by only 14.2\% (28.8→24.7) and SimCSE by just 1.4\% (88.1→86.9), revealing the robustness in multi-word precision and overall semantics, whereas BLEU‑1 and METEOR drop by 46.8\% (44.2→23.5) and 22.2\% (25.7→20.0), respectively, but mainly influence the single word similarity and fluency slightly. These results confirm that our model preserves semantic coherence to some extent under realistic adverse conditions. Figure~\ref{adverse} visualizes example outputs across the four scenarios to show our robustness.

\section{Conclusion}
In this work, we introduce RadarLLM, the first end-to-end LLM-based framework for semantic human motion understanding from mmWave point clouds, using physics-aware signal synthesis to create realistic radar–text pairs for addressing the data scarcity challenge. Experiments on synthetic data and real-world data in various conditions demonstrate state-of-the-art results. 
\subsubsection{Limitations and Future Work}
While RadarLLM demonstrates strong performance, several avenues remain for future exploration. First, although we calibrate simulation parameters to our TI AWR1843BOOST setup, evaluating and adapting RadarLLM across diverse radar hardware—with varying point‐cloud densities and range/angle resolutions—could further validate its generality. Second, extending our synthetic dataset to include environmental context and human–object interactions would enrich scene understanding and broaden applicability to more complex real‐world scenarios.

\appendix
\setcounter{page}{1}
\setcounter{figure}{0}
\setcounter{table}{0}
\setcounter{figure}{0}\renewcommand{\thefigure}{S\arabic{figure}}
\setcounter{table}{0}\renewcommand{\thetable}{S\arabic{table}}
\setcounter{equation}{0}\renewcommand{\theequation}{S\arabic{equation}}

\twocolumn[
	\begin{@twocolumnfalse}
		\section*{\centering{\Large{RadarLLM: Empowering Large Language Models to Understand Human Motion from Millimeter-wave Point Cloud Sequence}}}
		\begin{center}
			\large{Supplementary Material}
		\end{center}
		\centering
	\end{@twocolumnfalse}
]

\setcounter{section}{0}
\renewcommand\thesection{\Alph{section}}
\section{Dataset Details}
\subsection{IF Signal Simulation}
After the accumulation of simulated ray tracing paths, the received IF signal can be modeled as:

{\small
\begin{equation}
\label{eq1}
R_{IF}(t) = \sum_{i=0}^{N} A \exp\left( j  \frac{2\pi d}{c} \left( \frac{B}{T} t  + f_c  \right) \right),
\end{equation}}
where \( N \) is the number of rays, \( A \) is the antenna gain pattern, \( f_c \) is the carrier frequency, \( B \) is the signal bandwidth, \( T \) is the chirp duration, \( d \) is the ray path length and \( c \) is the speed of light.

To further bridge the gap between simulated and real-world radar signals, we introduce complex white Gaussian noise into the ideal signal \cite{2}:

{\small
\begin{equation}
\label{eq2}
R^{'}_{IF}(t) = R_{IF}(t) + \sqrt{{P_{\text{signal}}}/{10^{\text{SNR}/10}}} \cdot \epsilon(t),
\end{equation}}
where \( \epsilon(t) \sim \mathcal{N}(0, 1) \) is the standard Gaussian noise, \( P_{\text{signal}} = \mathbb{E}\left[|R_{IF}(t)|^2\right] \) is the power of the signal, SNR is the signal-to-noise ratio in dB, derived from real-world radar measurements. 

This ensures the synthesized signals maintain realistic noise-floor properties across varying motion intensities.

\subsection{Point Cloud Generation}
\label{sec: point cloud generation}
Initial spectral decomposition begins with dual Fourier transforms: the Range-FFT $\mathbf{R} = \mathcal{F}_r(R'_{IF})$ resolves radial distances through fast-time analysis, followed by Doppler-FFT $\mathbf{D} = \mathcal{F}_d(\mathbf{R})$ extracting velocity profiles along slow-time dimension. 

Static clutter suppression is performed across all $N_{rx}$ receive antennas:
{\small
\begin{equation}
    \bar{\mathbf{D}}^{(k)} = \mathbf{D}^{(k)} - \frac{1}{N_{rx}}\sum_{i=1}^{N_{rx}}\mathbf{D}_i^{(k)}, \quad k=1,...,N_{\text{frames}}
\end{equation}}
eliminating stationary reflections through spatial averaging. 

Adaptive peak selection ensures stable point density under varying SNR conditions:
{\small
\begin{equation}
    \mathcal{P} = \left\{(r_m, v_m, D_m)\right\}_{m=1}^{128} = \underset{(r,v)\in\mathbb{Z}^2}{\text{argtopk}_{128}} \left(|\bar{\mathbf{D}}|\right),
\end{equation}}
where $D_m = \bar{\mathbf{D}}(r_m, v_m)$ retains complex Doppler-FFT values.

Physical decoding generates spatial coordinates and derived parameters:
{\small
\begin{align}
    r &= \frac{c \cdot r_m}{2B}, \quad \theta = \arcsin\left(\frac{\lambda v_m}{2d_{\text{max}}}\right), \\
    \phi &= \arctan\left(\frac{y_{\text{ant}}}{x_{\text{ant}}}\right), \\
    x &= r \sin\theta \cos\phi, \quad y = r \sin\theta \sin\phi, \\
    z &= r \cos\theta, \quad v = \frac{\lambda v_m}{2},
\end{align}}
where $B$ denotes signal bandwidth and $(x_{\text{ant}}, y_{\text{ant}})$ specifies antenna array geometry.

The final 6D feature vector integrates spatial-temporal and spectral attributes:
{\small
\begin{equation}
    \mathbf{p}_m = \left[x, y, z, r, v, 10\log_{10}(|D_m|)\right]^T \in \mathbb{R}^6.
\end{equation}}
The 6D features are 3D Cartesian coordinates $(x,y,z)$, radial distance $r$, velocity $v$, and log-scaled Doppler intensity, respectively. This representation preserves electromagnetic scattering characteristics while maintaining compatibility with neural feature extractors.

\begin{figure}
    \centering
    \includegraphics[width=1\linewidth]{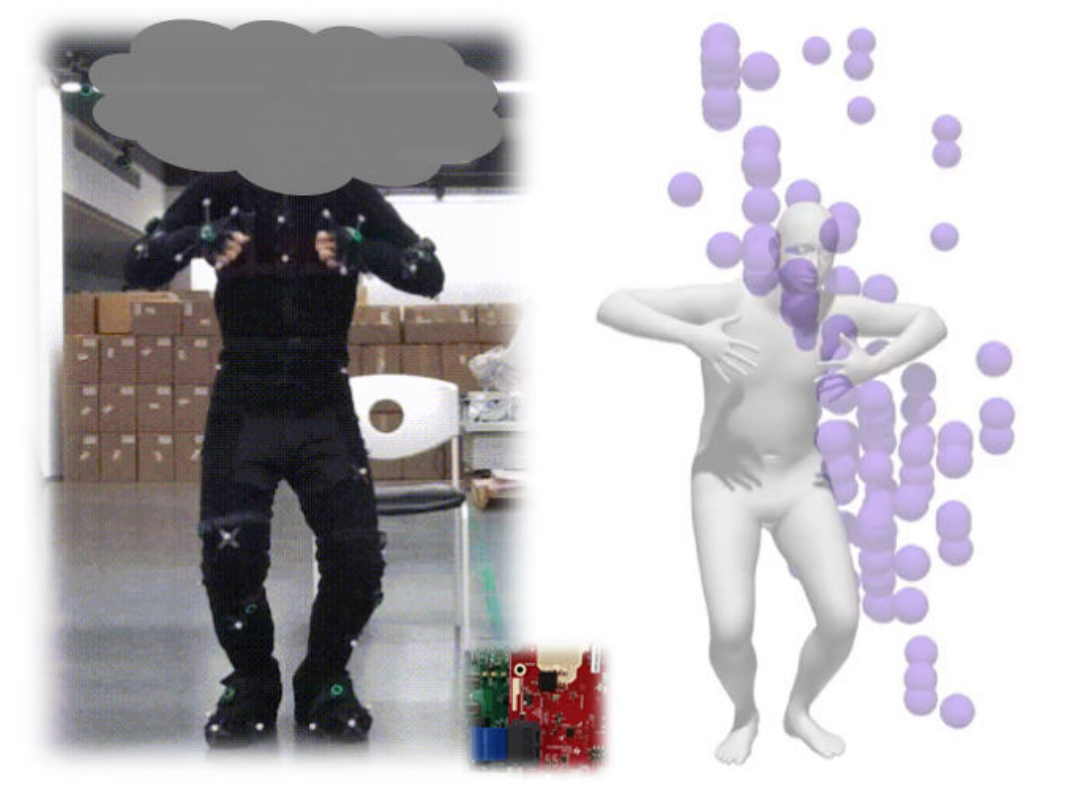}
    \caption{Real dataset setup.}
    \label{captionreal}
\end{figure}

\subsection{Real Dataset}
\subsubsection{Hardware implementation.}The millimeter-wave radar we used in this paper is TI AWR1843BOOST, and the TI DCA1000EVM is connected to enable data capture and streaming from mmWave radar. This mmWave device contains 3 transmitting antennas and 4 receiving antennas. For each FMCW chirp, the frequency increases from 77 GHz to 80.9 GHz, and each chirp is composed of 256 sampling points. The mmWave device is set to send 10 frames per second, and each frame is composed of 128 chirps. Based on our device setting, the maximum sensing range of the mmWave device is about 11 m, the range resolution is about 4.3 cm, the maximum sensing velocity is about 4.5 m/s, and the velocity resolution is about 7.1 cm/s.
\par

\subsubsection{Dataset Collection and Annotation.}
The collected dataset consists of 125 different motions from the HumanML3D test set. Our collection setup is shown in Figure~\ref{captionreal}. The subject is instructed to mimic each motion and repeat it three times, resulting in a total of 375 motion sequences. The duration of each sequence varies between 6 to 9 seconds.
Motion descriptions are based on the original text annotations and have been carefully reviewed to ensure better alignment with the actual motions. The motions take place in a $5\text{m} \times 5\text{m}$ area, with the subject positioned 3 meters away from the radar at the start of each sequence to minimize location-based influence.
To enhance visualization and enable future research applications, an Azure Kinect RGB-D camera and the OptiTrack motion capture (MoCap) system are also used during data collection. Additionally, the captured motions are processed into the SMPL-X format using Mosh++ \cite{mahmood2019amass}.

{\small
\begin{algorithm}[t]
\caption{RadarLLM training and inference}
\label{alg:radarllm}
\SetKwInOut{Input}{Input}
\SetKwInOut{Output}{Output}
\SetKwInOut{Initialize}{Initialize}

\Input{Virtual training dataset $\mathcal{D}_{\text{train}} = \{\mathbf{P}_i, \mathbf{m}_i, \mathbf{Y}_i, \mathcal{I}_i\}_{i=1}^T$, Virtual test dataset $\mathcal{D}_{\text{test-vir}} = \{\mathbf{P}_i, \mathbf{m}_i, \mathbf{Y}_i, \mathcal{I}_i\}_{i=1}^T$, Real dataset $\mathcal{D}_{\text{test-real}} = \{\mathbf{P}_i, \mathbf{Y}_i^*, \mathcal{I}_i\}_{i=1}^{T'}$}
\Output{Motion description predictions $\mathbf{\hat{Y}}$}
\Initialize{Aggregate VQ-VAE encoder $\mathbf{E}_\phi$, decoder $\mathbf{D}_\psi$, codebook $\mathcal{Z}$, T5 model $\mathcal{M}_\omega^*$ with extended vocabulary $\mathcal{V}$}

\BlankLine
\SetKwBlock{Stage1}{Stage 1: Radar Tokenizer Training}{}
{
    \For{epoch = 1 to $E_{\text{token}}$}{
        \For{batch $\mathbf{P} \sim \mathcal{D}_{\text{train}}$}{
            $\mathbf{F}_{\text{vis}}, \mathbf{F}_{\text{msk}} \leftarrow \textit{Template Prior Grouping}(\mathbf{P})$\;
            $\mathbf{F}_{\text{all}} \leftarrow \mathbf{F}_{\text{vis}} \cup \mathbf{D}_\psi(\mathbf{F}_{\text{vis}})$\;
            $\mathbf{z} \leftarrow \text{Quantize}(\mathbf{F}_{\text{all}}, \mathcal{Z})$\;
            Compute $\mathcal{L}_{\text{rec}}$, $\mathcal{L}_{\text{emb}}$, $\mathcal{L}_{\text{commit}}$ via Eq.(5-7)\;
            $\phi,\psi \leftarrow \text{AdamW}(\nabla_{\phi, \psi}(\mathcal{L}_{\text{rec}} + \mathcal{L}_{\text{emb}} + \mathcal{L}_{\text{commit}}))$\;
        }
    }
}

\BlankLine
\SetKwBlock{Stage2}{Stage 2: Radar-Language Pretraining}{}
{
    \For{epoch = 1 to $E_{\text{pretrain}}$}{
        \For{batch $(\mathbf{P}, \mathbf{Y}) \sim \mathcal{D}_{\text{train}}$}{
            $\mathbf{z} \leftarrow \mathbf{E}_\phi(\mathbf{P})$\;
            Compute $\mathcal{L}_{\text{pred}}$ via Eq. 8\;
            Compute $\mathcal{L}_{\text{r2t}}$, $\mathcal{L}_{\text{t2r}}$ via Eq. (9,10)\;
            $\omega \leftarrow \text{AdamW}(\nabla_\omega(\lambda_1\mathcal{L}_{\text{pred}} + \lambda_2\mathcal{L}_{\text{r2t}} + \lambda_3\mathcal{L}_{\text{t2r}}))$\;
        }
    }
}

\BlankLine
\SetKwBlock{Stage3}{Stage 3: Instruction-Aware Fine-Tuning}{}
{
    \For{epoch = 1 to $E_{\text{tune}}$}{
        \For{batch $(\mathcal{I}, \mathbf{P}, \mathbf{Y}) \sim \mathcal{D}_{\text{train}}$}{
            $\mathbf{X} \leftarrow \text{Concat}(\mathcal{I}, \mathbf{E}_\phi(\mathbf{P}))$\;
            Compute $\mathcal{L}_{\text{tune}} = -\sum \log p(\mathbf{Y}|\mathbf{X})$\;
            $\omega \leftarrow \text{AdamW}(\nabla_\omega\mathcal{L}_{\text{tune}})$\;
        }
    }
}

\BlankLine
\SetKwBlock{Stage4}{Stage 4: Inference}{}
{
    \For{each $\mathbf{P} \in [\mathcal{D}_{\text{test-vir}}, \mathcal{D}_{\text{test-real}}]$}{
        $\mathbf{\hat{Y}} \leftarrow \mathcal{M}_\omega^*(\mathbf{E}_\phi(\mathbf{P}))$\;
    }
    \Return $\mathbf{\hat{Y}}$\;
}
\end{algorithm}}

\section{Algorithms}
The following algorithm outlines the RadarLLM training and inference pipeline, comprising four stages: (1) Radar Tokenizer Training for encoding radar sequences into tokens, (2) Radar-Language Pretraining to align radar and text tokens, (3) Instruction-Aware Fine-Tuning for radar-to-text adaptation, and (4) Inference on virtual and real test data.

\section{Implementation Details}
To ensure reproducible results, we implement \textbf{RadarLLM} using PyTorch 2.0. The radar tokenizer employs an encoder $\mathbf{E}$ with a temporal down-sampling rate $l$ derived from temporal stride 2 and token unit 2, paired with a codebook $\mathcal{K} \in \mathbb{R}^{512 \times 512}$ (optimal size validated in Table~\ref{tab:codebook_size}). For the language model, we initialize a FLAN-T5-small backbone with 6-layer transformer encoder/decoders, where feed-forward networks adopt 2048-dimensional outputs while other sub-layers maintain 512-dimensional embeddings. All models utilize the AdamW optimizer with distinct learning rates: $3.5 \times 10^{-4}$ for radar tokenizer training (100 epochs), and $2 \times 10^{-4}$ for language model pretraining (300 epochs) and instruction tuning (100 epochs). Training employs a unified batch size of 16 across all stages, with mixed radar-text batches enabling joint unsupervised and supervised learning. The entire framework is executed on an RTX3090 GPU.

\section{More Qualitative Evaluation}
More qualitative results are presented alongside the action labels predicted by RadHAR~\cite{singh2019radhar}. On the one side, our method consistently outperforms both motion- and video-based baselines in generating accurate motion descriptions and contextual understandings. On the other side,  while the action labels predicted by the HAR method may partially reflect the overall action tendency, they lack fine-grained motion details and fail to distinguish between different actions over time.

\begin{figure*}[htbp!]
    \centering
    \includegraphics[width=\linewidth]{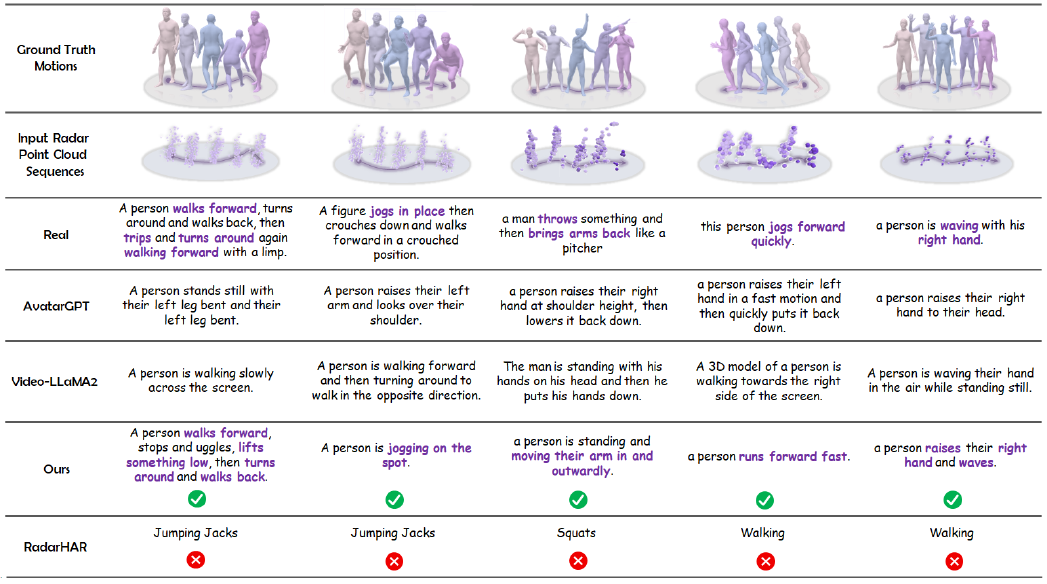}
    \caption{Additional visualization of prediction text and corresponding motion and radar points. The left two are tested on real data, while the right three are on virtual data.}
    \label{fig:blank}
\end{figure*}

\setlength{\tabcolsep}{1.1mm}
\begin{table}[t]
\centering
\fontsize{9}{10}\selectfont
\begin{tabular}{lcccc}
\toprule[1pt]
Dataset & MPJRE (◦) & MPJPE (cm) & MPVPE (cm) & MTE (cm) \\
\midrule
mmBody & 22.08 & 28.91 & 21.60 & 15.46 \\
MRI & 9.94 & 11.55 & - & 8.25 \\
Ours & 19.74 & 15.87 & 18.70 & 11.27 \\
\bottomrule[1pt]
\end{tabular}
\caption{HPE metrics of mmMesh on different datasets.}
\label{tab:hpe}
\end{table}

\setlength{\tabcolsep}{0.82mm}
\begin{table}[t]
\centering
\fontsize{9}{10}\selectfont
\begin{tabular}{cccccc}
\toprule[1pt]
\makecell[c]{Codebook \\ Size}  & ROUGE-L & BLEU-1  & METEOR & CIDEr & BERTScore  \\
\midrule
256  & 25.6 & 29.3  & 23.2 & 29.0 & 80.8  \\
512   & \textbf{34.9} & \textbf{46.4}  & \textbf{33.2} & \textbf{5.9} & \textbf{83.0}  \\
1024  & 24.2 & 34.0  & 22.2 & 3.5 & 81.7  \\
2048  & 15.2 & 20.4  & 14.4 & 2.6 & 78.6  \\
\bottomrule[1pt]

\end{tabular}
\caption{Ablation study on codebook size.}
\label{tab:codebook_size}
\end{table}
\section{Evaluation on Radar Point Cloud-based Human Pose Estimation}
To further evaluate the quality of the virtual dataset and demonstrate that the human motions generated by mmMesh are reasonable inputs for the baseline methods, we conduct comparisons on the mmBody \cite{chen2022mmbody} and MRI \cite{an2022mri} datasets. Following \cite{27}, we use Mean Per Joint Rotation Error (MPJRE), Mean Per Joint Position Error (MPJPE), Mean Per Vertice Position Error (MPVPE) and Mean Translation Error (MTE) as the evaluation metrics. The results from the virtual test dataset are similar to those of the mmBody dataset, while a performance gap is observed with MRI dataset. It arises because the MRI dataset primarily focuses on simple rehabilitation movements. 

\section{Ablation Study of Codebook Size}
As shown in Table~\ref{tab:codebook_size}, we conduct experiments with different codebook sizes ranging from 256 to 2048. Among them, a codebook size of 512 achieves the best performance across all metrics. A smaller codebook size (256) restricts the model’s expressive capacity, while larger codebooks (1024 and 2048) introduce quantization instability. This effect is further evidenced by the performance degradation from 1024 to 2048.

\section{Ablation Study of instruction tuning}
\begin{figure}[t!]
    \centering
    \includegraphics[width=\linewidth]{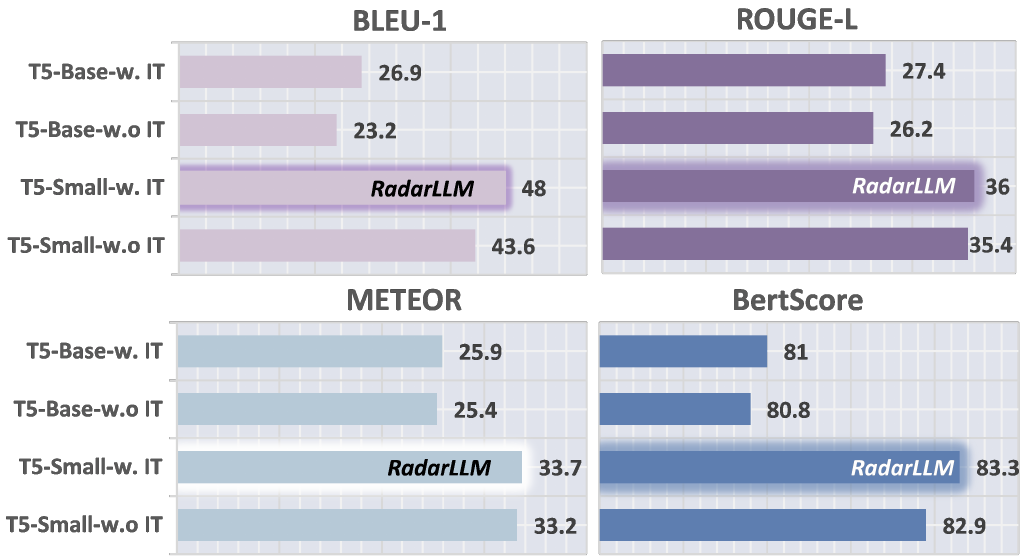}
    \caption{Ablation study on LLM instruction tuning.}
    \label{fig:insllm}
\end{figure}

We conduct comprehensive instruction tuning experiments to validate the efficacy of our multi-modal adaptation strategy. As shown in Figure~\ref{fig:insllm}, instruction tuning (T5-Small w.IT) significantly improves the performance of T5-Small, with a 15.9\% increase in BLEU-1, demonstrating enhanced task-specific knowledge integration. However, T5-Base model with instruction tuning (T5-Base w.IT) shows diminishing returns with only a 10.09\% improvement in BLEU-1, suggesting that the current radar token representation may not fully utilize the model's capacity. This is because T5-Small has fewer parameters, it can learn radar-text alignment during pretraining more effectively, allowing instruction tuning to further refine its textual understanding and boost the performance.


\section{Ablation Study of Vector quantization necessity}
Ablation in Table~\ref{2} shows that incorporating VQ yields a 1.4\% increase in ROUGE-L, a 4.7\% boost in BLEU-1, and a 1.2\% gain in BERTScore, at the expense of a minor 1.5\% drop in METEOR. This demonstrates that VQ’s discrete codebook effectively filters out radar noise and redundant variations, sharpening core motion features and improving cross-modal alignment, with a slight METEOR decrease for larger lexical accuracy.

\setlength{\tabcolsep}{1.8mm}
\begin{table}[t]
  \centering
  \fontsize{9}{10}\selectfont
  \begin{tabular}{lcccc}
    \toprule
    Setting        & ROUGE-L & BLEU-1 & METEOR & BERTScore \\
    \midrule
    w/o VQ         & 29.8    & 34.3   & \textbf{31.7} & 81.0      \\
    \textbf{w/ VQ} & \textbf{31.2} & \textbf{39.0} & 30.2          & \textbf{82.2} \\
    \bottomrule
  \end{tabular}
  \caption{Ablation of vector quantization with model only pre-trained on radar-to-text.}
  \label{2}
\end{table}

\section*{Acknowledgements}
This work was supported by the National Natural Science Foundation of China (NSFC) under Grant 62273229.

{\small
\bibliography{main}
}



\end{document}